\documentclass{article}
\usepackage{arxiv}

\usepackage{times}
\usepackage{soul}
\usepackage{url}
\usepackage{hyperref}
\usepackage[utf8]{inputenc}
\usepackage[small]{caption}
\usepackage{graphicx}
\usepackage{amsmath}
\usepackage{amsthm}
\usepackage{booktabs}
\usepackage{algorithm}
\usepackage{algorithmic}
\usepackage{subcaption}
\usepackage{enumitem}
\usepackage{multirow}
\usepackage{cite}

\usepackage{amsmath,amsfonts,bm}

\def\eqref#1{equation~\ref{#1}}
\def\Eqref#1{Equation~\ref{#1}}

\def\1{\bm{1}}

\def\vx{{\bm{x}}}
\def\vy{{\bm{y}}}
\def\vz{{\bm{z}}}

\def\evx{{x}}
\def\evy{{y}}

\def\mA{{\bm{A}}}
\def\mB{{\bm{B}}}

\def\mE{{\bm{E}}}

\def\mI{{\bm{I}}}

\def\mP{{\bm{P}}}

\def\mX{{\bm{X}}}
\def\mY{{\bm{Y}}}

\DeclareMathAlphabet{\mathsfit}{\encodingdefault}{\sfdefault}{m}{sl}
\SetMathAlphabet{\mathsfit}{bold}{\encodingdefault}{\sfdefault}{bx}{n}

\def\gG{{\mathcal{G}}}

\def\sV{{\mathbb{V}}}

\def\emE{{E}}
\def\emF{{F}}

\DeclareMathOperator*{\argmin}{arg\,min}

\newcommand{\angstrom}{\mbox{\normalfont\AA}}

\title{Learning Geometrically Disentangled Representations of Protein Folding Simulations}

\author{
N. Joseph Tatro
\thanks{Currently at Systems \& Technology Research (STR) in Woburn, MA} \\
Department of Mathematical Sciences \\
Rensselaer Polytechnic Institute \\
Troy, NY 12180 \\
\texttt{tatron@rpi.edu} \\
\And 
Payel Das, Pin-Yu Chen, \& Vijil Chenthamarakshan \\
IBM Research \\
Yorktown, NY 10598 \\
\AND 
Rongjie Lai \\
Department of Mathematical Sciences \\
Rensselaer Polytechnic Institute \\
Troy, NY 12180
}

\begin{document}

\maketitle

\begin{abstract}
Massive molecular  simulations of drug-target proteins have been used as a tool to understand disease mechanism and develop therapeutics. This work focuses on learning a generative neural network on a structural ensemble of a drug-target protein, e.g. SARS-CoV-2 Spike protein, obtained from computationally expensive molecular simulations. Model tasks involve characterizing the distinct structural fluctuations  of the protein bound to various drug molecules, as well as efficient generation of  protein conformations that can serve as an complement of a molecular simulation engine. Specifically, we present a  geometric autoencoder framework to learn separate latent space encodings of  the intrinsic and extrinsic geometries of the protein structure. For this purpose, the proposed Protein Geometric AutoEncoder (ProGAE) model is trained on  the protein contact map and the  orientation of the backbone bonds of the protein. Using ProGAE latent embeddings, we reconstruct and generate the conformational ensemble of a protein at or near the experimental resolution, while gaining better interpretability and controllability in term of protein structure generation from the learned latent space. Additionally, ProGAE models are transferable to a different state of the same protein or to a new protein of different size, where only the dense layer decoding from the latent representation needs to be retrained. Results show that our geometric learning-based method enjoys both accuracy and efficiency for generating complex structural variations, charting the path toward scalable and improved approaches for analyzing and enhancing high-cost simulations of drug-target proteins.
\end{abstract}

\section{Introduction}

\begin{figure*}[t]
    \centering
    \includegraphics[width=0.75\textwidth]{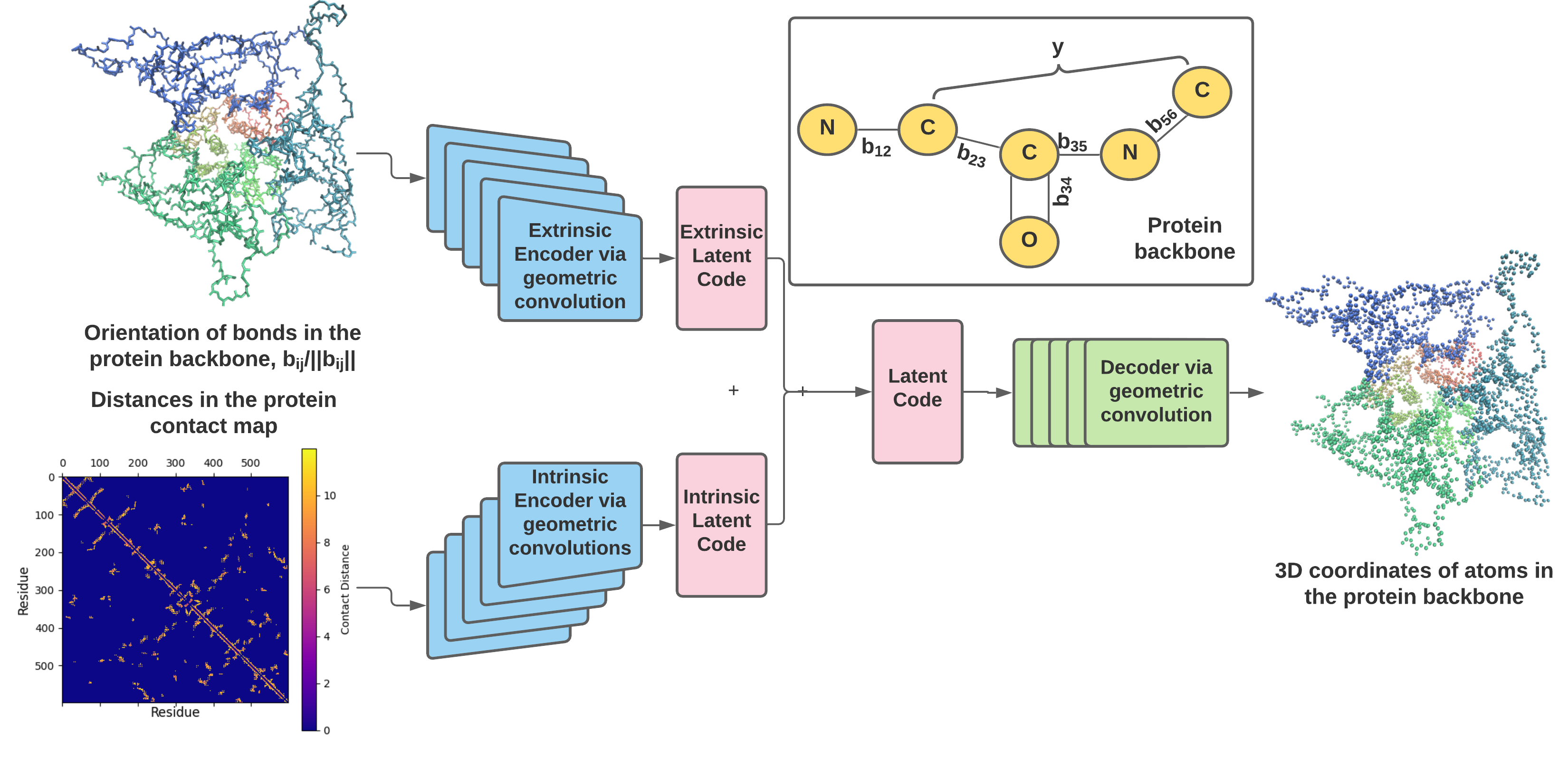}
    \caption{Architecture of ProGAE, it generates protein conformations via separate encoding of data related to coarse intrinsic and extrinsic geometries. These geometries are captured via the orientation of the backbone bonds (extrinsic) and distances of close contacts of alpha carbons (intrinsic), as captured by the contact map. These latent representations are jointly used to generate the 3D coordinates of the backbone atoms.}
    \label{fig:autoencoder_arch}
\end{figure*}

\begin{figure}[tb]
\centering
    \begin{subfigure}[b]{0.48\linewidth}
         \centering
         \includegraphics[width=\textwidth]{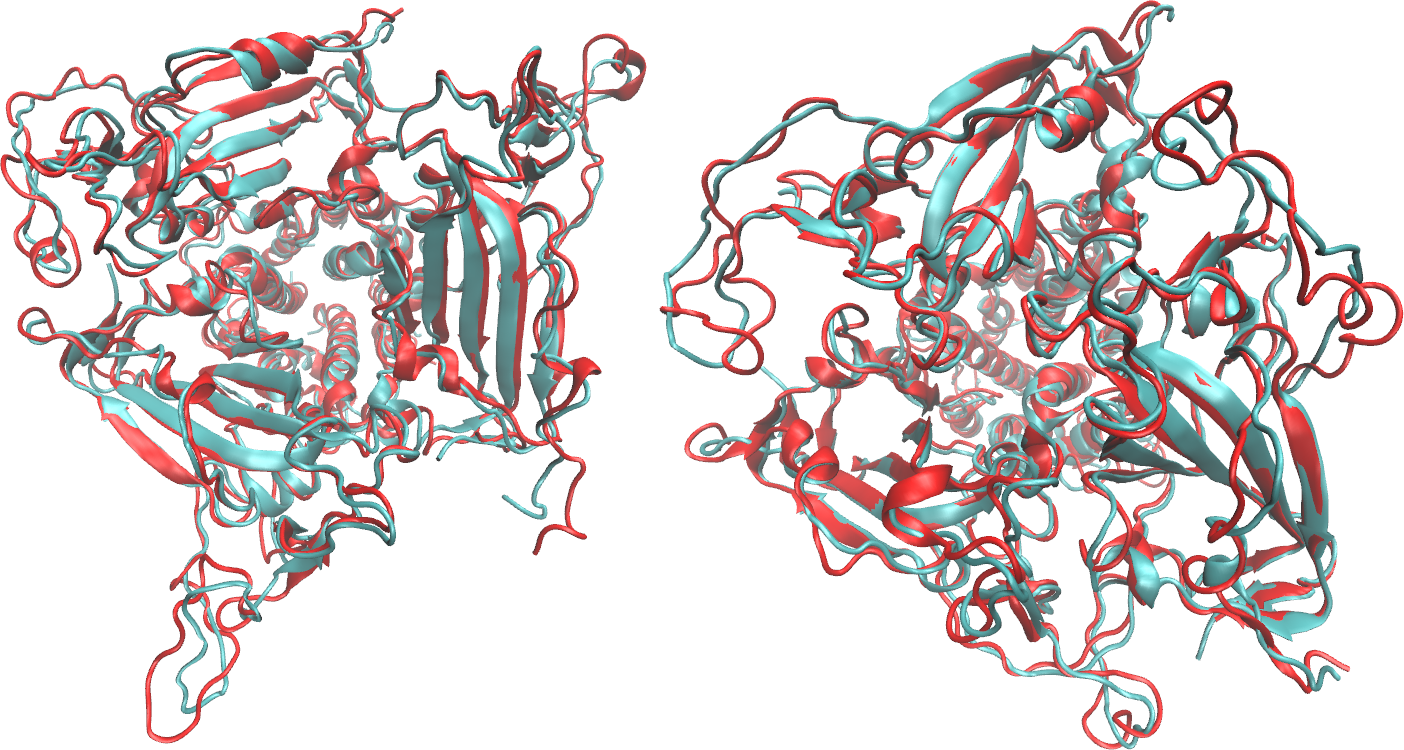}
     \end{subfigure}
     \begin{subfigure}[b]{0.48\linewidth}
         \centering
         \includegraphics[width=\textwidth]{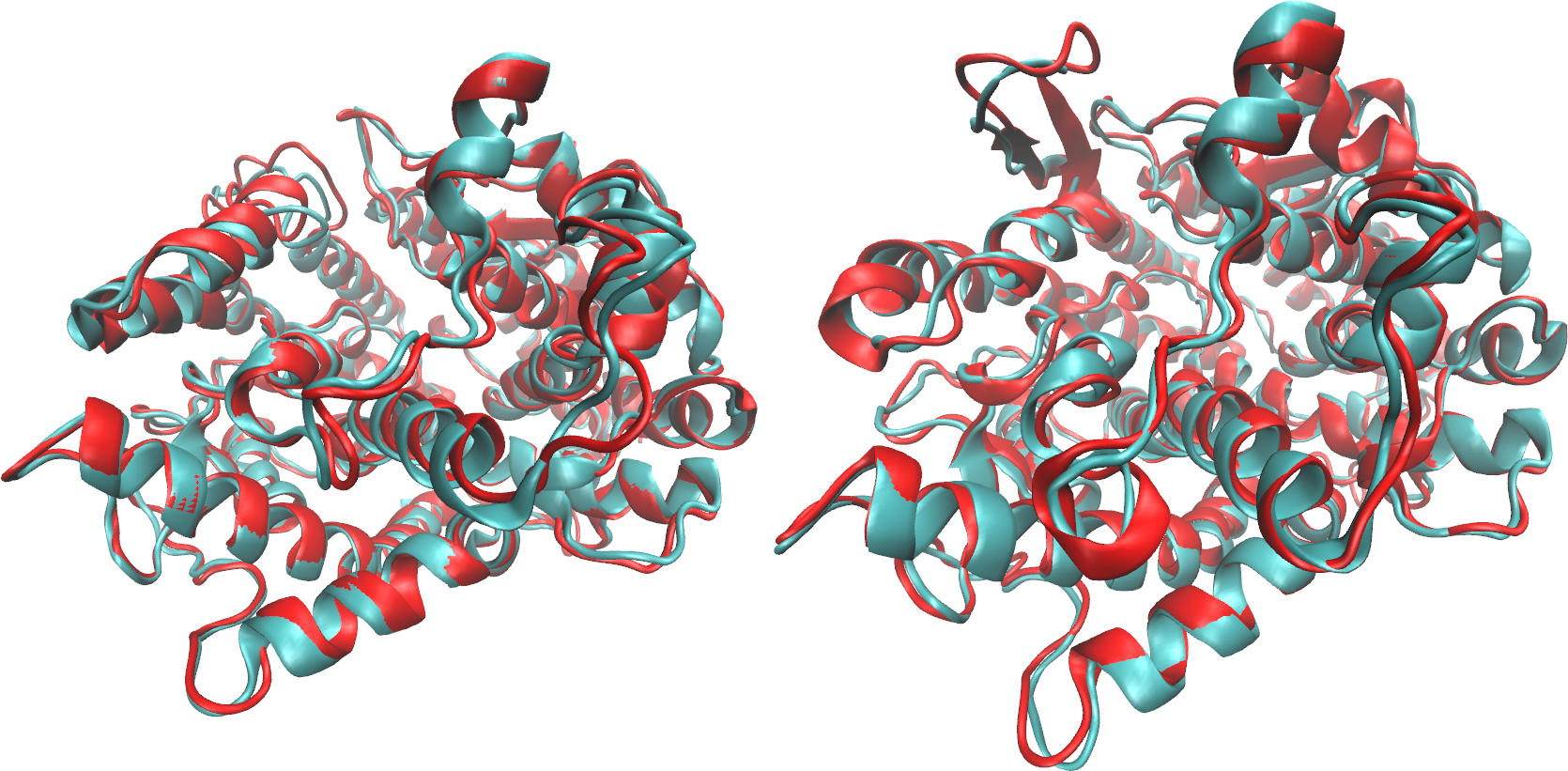}
     \end{subfigure}
\caption{ProGAE reconstructions of S protein in SARS-CoV-2 (left) and hACE2 data (right). Blue and red structures correspond to the reconstructed and ground truth structures, respectively, and are superimposed. The RMSD associated with the visualized structures are $1.57 \angstrom, 1.65 \angstrom, 0.96 \angstrom,$ and $1.08 \angstrom$. 
}
\label{fig:recon}
\end{figure}


Understanding the protein conformational landscape is critical, as protein functions
are intimately connected with structural variations. Recently, deep learning based models have made impressive progress in accurately predicting protein structures~\cite{jumper2021highly}. 
There has been also interest in modeling the underlying conformational space of proteins by using deep generative models, e.g.  
\cite{ramaswamy2020learning} 
and 
\cite{ bhowmik2018deep,guo2020generating,varolgunecs2020interpretable}.
This line of  work has mainly attempted to respect the domain geometry by using convolutional AEs on features extracted from 3D structures. In parallel, learning directly from 3D structure has recently developed into an exciting and promising application area for deep learning. 

In this work, we learn a model of the protein conformational space from a set of protein simulations by using
geometric deep learning. We  also investigate  how the geometry of a protein itself can assist learning and improve interpretability of the latent conformational space. Namely, we consider the distinct influence of the  intrinsic and extrinsic geometries. 
The intrinsic geometry captures the global (slow timescale) structural information of a protein, whereas the extrinsic geometry accounts for the local (fast timescale) structural variations. 
The  structural variations can be further induced due to interactions with the external environment (drug molecule in the current case). 
Intrinsic geometric properties can be thought to be more robust to minor protein conformational change. 
To this end, we propose a Protein Geometric Autoencoder model, named ProGAE, to separately encode intrinsic and extrinsic protein geometries and subject that to understanding the structural landscape of established drug-target proteins, such as COVID-19  target proteins and GPCRs.     


The main contributions of this work are summarized: 
\begin{itemize}[leftmargin=*]
\item Inspired by recent unsupervised geometric disentanglement learning  works \cite{tatro2020unsupervised,wu2019disentangling, yang2020dsmnet}, we propose a novel geometric autoencoder named ProGAE that directly learns from 3D protein structures via separately encoding intrinsic and extrinsic geometries into disjoint latent spaces used to generate protein structures.
We propose a novel formulation, in which network intrinsic input is taken as the protein contact map, and the extrinsic input is the backbone bond orientations. 
\item We find that the intrinsic geometric latent space improves the quality of the reconstructed proteins. Experiments confirm that the extrinsic geometry of proteins generated by ProGAE accounts for the structural variations due to specific drug interaction. This allows for better latent space interpretability and  controllable generation. 
\item Analysis shows the learned extrinsic geometric latent space can be used for drug classification and drug property prediction, where the drug is bound to the given protein. We also demonstrate that the learned ProGAE can be transferred to a trajectory of the protein in a different state or a trajectory of a different protein all-together.
\end{itemize}

\subsection{Related Work}

Recently, a body of work has used deep learning to learn from protein structures \cite{graves2020review, jing2020learning}. For example, \cite{gainza2019deciphering} uses geometric deep learning to predict docking sites for protein interactions. \cite{hermosilla2020proteinn} leverages the notion  of intrinsic and extrinsic geometry to define an architecture for a fold classification task.

Additionally, there has been focus on directly  learning the temporal aspects of molecular dynamics from simulation trajectories, which is not directly related to the current work. Please see Appendix \ref{app:add_back} for a detailed discussion. 
Several recent papers use AE-based approaches for either analyzing and/or generating structures from  the  latent space \cite{bhowmik2018deep, guo2020generating, ramaswamy2020learning, varolgunecs2020interpretable}, which are most closely related to this work. \cite{bhowmik2018deep} and \cite{guo2020generating} aim at learning from and generating protein contact maps, while ProGAE directly deals with 3D structures. Therefore a direct comparison of ProGAE with these methods is not possible. \cite{varolgunecs2020interpretable} uses a VAE with a Gaussian Mixture Prior for performing clustering of high-dimensional input configurations in the learned latent space. While the method works well on toy models and a standard Alanine Dipeptide benchmark, its performance drops as the size of the protein system  grows to 15 amino acids, which is approximately an order smaller than  the protein systems studied here. \cite{ramaswamy2019learning} trains a 1D CNN autoencoder on backbone (including beta carbon) coordinates and uses a loss objective comprised of geometric MSE error and physics-based (bond length, bond angle, nonbonded) error. Due to the unavailability of code or pre-trained model, we were unable to perform a direct comparison. Nevertheless, we run ProGAE on the same MurD protein simulations studied in  \cite{ramaswamy2019learning} and compare the  reconstruction quality with respect to the value reported in that study as  well as to the experimental resolution. 

None of these  works has  considered explicit disentangling of intrinsic and extrinsic geometries. 
To the best of our knowledge, this work is the first to propose an autoencoder for the unsupervised modeling of the geometric disentanglement of protein conformational space captured in molecular simulations. This representation provides better interpretability of the latent space, in terms of the physico-chemical and geometric attributes, results in more geometrically accurate protein conformations, as well as scales and transfers well to larger protein systems.

\section{ProGAE for Protein Conformational Space}

\begin{figure}[t]
\centering
\begin{subfigure}[b]{0.75\textwidth}
     \centering
     \includegraphics[width=\textwidth]{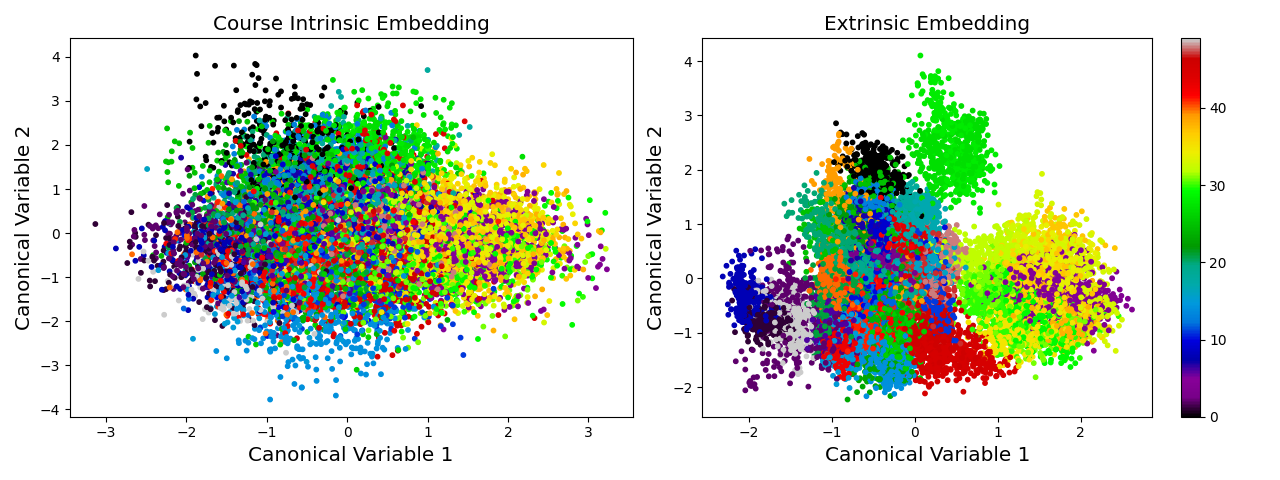}
     \caption{S protein}
     \label{fig:cca_sarscov2}
 \end{subfigure}
\caption{The projection of the latent space embedding to the first two canonical vectors between the intrinsic and extrinsic latent spaces. Color indicates the identity of the drug that the protein is bound to in that conformation. Clustering by drug identity is apparent in the extrinsic latent space, but much weaker in the intrinsic latent space, consistent with Table \ref{tab:cca_results}.}
\end{figure}

\begin{small}
\begin{table*}[t]
\centering
\begin{tabular}{@{}rrrcrcrcr@{}}
\toprule
& Errors & \textbf{S protein} & & \textbf{hACE2} & & \textbf{MurD} & & \textbf{GPCR}    \\
\midrule 
& \textit{Exp. Res. (\angstrom)} & 2.68/2.80 && 2.20/3.00 && 2.40/1.77/1.84 && - \\
\cmidrule{1-2}
\textit{Training} & Loss (E-1) & $9.12 \pm 0.09$ && $4.54 \pm 0.0.$ && $3.04 \pm 0.01$ && $8.16 \pm 0.11$  \\
& Avg. $L_2$ (\angstrom) & $1.32 \pm 0.01$ && $0.82 \pm 0.0.$ && $0.64 \pm .00$ && $1.21 \pm 0.01$\\
& Contact Recov. & $70.6 \pm 0.1 \%$ && $78.7 \pm 0.1 \%$ && $86.0 \pm 0.3 \%$ && $80.3 \pm 0.2 \%$ \\
\cmidrule{1-2}
\textit{Test} & Loss (E-1) & $9.42 \pm 0.08$  && $4.68 \pm 0.00$ && $6.05 \pm 0.06$ && $8.22 \pm 0.08$ \\
& Avg. $L_2$ (\angstrom) & $1.35 \pm 0.01$ && $0.83 \pm 0.00$ && $0.98 \pm 0.01$ && $1.22 \pm 0.01$ \\
& Contact Recov. & $70.4 \pm 0.1 \%$ && $78.5 \pm 0.0 \%$ && $84.2 \pm 0.2 \%$ && $80.2 \pm 0.01 \%$ \\ 
& RMSD (\angstrom) & $1.76 \pm 0.18$ && $1.00 \pm 0.09$ && $1.00 \pm 0.15$ &&  - \\
& SASA (\angstrom) & $-6.8 \pm 0.1 \%$ && - && - && - \\
\bottomrule
\end{tabular}
\caption{Average atom-wise $L_2$ error on the training/test sets, as well as RMSD between reconstructed and true structures, using ProGAE. The Jaccard index of the reconstructed binarized contact map with a cutoff of $8 \angstrom$ is provided. The RMSD on test set are within the resolution of the associated PDB files. We also show the relative error in surface-accessible surface area of the binding pocket of the S protein.}
\label{tab:reconstruction_results}
\end{table*}
\end{small}

\begin{table}[t]
\centering
\begin{tabular}{@{}rrcr@{}}
\toprule
Drug classification & \textbf{S protein} & \phantom{a} & \textbf{hACE2}\\
\midrule 
1\textsuperscript{st} canonical corr. & $0.80 \pm 0.00$ && $0.87 \pm 0.01$ \\
Trained on intrinsic & $53.8 \pm 0.1 \%$ && $51.0 \pm 0.0$ \% \\
Trained on extrinsic & $94.1 \pm 0.3 \%$ && $94.2 \pm 0.2$ \% \\
\bottomrule
\end{tabular}
\caption{The leading canonical correlation between the learned intrinsic and extrinsic latent spaces, as well as the performance of a linear model trained on disentangled latent spaces of ProGAE for drug classification. The linear model is trained with one-shot learning.
}
\label{tab:cca_results}
\end{table}

\begin{small}
\begin{table*}[t]
\centering
\begin{tabular}{@{}rrrcrcr@{}}
\toprule
\multirow{2}{*}{Dataset} & & \multirow{2}{*}{\shortstack[l]{Molecular \\ weight}} & \phantom{ab} & \multirow{2}{*}{\shortstack[l]{Hydrogen bond \\ donor count}} & \phantom{ab} & \multirow{2}{*}{\shortstack[l]{Topological polar \\ surface area}} \\
\\
\midrule 
S protein & PCA error ($\sigma$) & $0.71 \pm 0.00$ && $0.76 \pm 0.01$ && $0.74 \pm 0.00$ \\
& Latent error ($\sigma$) & $\mathbf{0.51 \pm 0.01}$ && $\mathbf{0.51 \pm 0.01}$ && $\mathbf{0.52 \pm 0.03}$ \\
\cmidrule{1-2}
hACE2 & PCA error ($\sigma$) & $0.67 \pm 0.00$ && $0.61 \pm 0.00$ && $0.69 \pm 0.00$ \\
& Latent error ($\sigma$) & $\mathbf{0.54 \pm 0.01}$ && $\mathbf{0.54 \pm 0.02}$ && $\mathbf{0.54 \pm 0.01}$ \\
\bottomrule
\end{tabular}
\caption{Results of linear regression on the extrinsic latent space for predicting physical and chemical properties of the drugs that a protein is bound to. Error is normalized for interpretability. For comparison, performance of linear regression on the PCA embeddings of the orientation of the backbone bonds is reported. This embedding is restrained to the same dimension as the latent space. 
}
\label{tab:classifier_results}
\end{table*}
\end{small}

First, we introduce the input signals for our novel geometric autoencoder, ProGAE. We then discuss how ProGAE utilizes these to generate the disentangled space. 

\paragraph{Geometric Features of Protein as Network Input}

ProGAE separately encodes intrinsic and extrinsic geometry with the goal of achieving better latent space interpretability. We clarify these geometric notions. Mathematically, we can consider a manifold (i.e. surface) independent of its embedding in Euclidean space. Properties that do not depend on an embedding are known as intrinsic geometric properties, with others referred to as extrinsic. As an example, suppose we approximate a protein structure via a graph, $\mathcal{G}$, where certain atoms are connected by edges. Then, the lengths of edges in this graph are intrinsic, as they are not explicitly dependent on the 3D embedding of said graph. On the other hand, the orientations of the edges of the embedded graph are extrinsic. 


As we will train ProGAE to learn the conformational space of a given protein, the protein primary structure is implicit. Then in treating it geometrically, we view the protein at the level of its backbone, which specifies its shape. 
Given primary structure, reconstructing this backbone is sufficient for reconstructing the detailed protein structure. Of importance in the backbone are the $C_\alpha$ atoms, which are the centers of amino acids in the protein. A coarse characterization of the backbone is the protein contact map, an incomplete distance matrix between all $C_\alpha$ atoms that contains all distances less than a specified threshold, typically between $6.5 \angstrom$ and $12 \angstrom$ \cite{vendruscolo1997recovery}. As a thresholded distance matrix, the contact map defines a graph structure on a protein conformation that we will refer to as the contact graph. For our purposes, the contact graph will exclude edges between residues $i$ and $i+j$ for $j \leq 3$. We will use the backbone and contact graph as domains for defining our signals. 

Both the protein backbone and contact graph can be viewed as polygonal chain in Euclidean space. They are depicted in Figure \ref{fig:autoencoder_arch} with their geometric features as network input. We see that a polygonal chain is determined up to translation given both the length and orientation of its line segments. Then it follows that the backbone can be determined given the length and orientation of its bonds. Here the length of these bonds is intrinsic while the orientation is extrinsic. Thus, to explicitly decouple the intrinsic and extrinsic geometry, we consider separately encoding these signals. 

We note that the length of covalent bonds undergo very little change during a simulation performed using an empirical force-field, like those in this work. As a result, a standard deviation of less than $0.059 \angstrom$ from target bond lengths is common in PDB structures \cite{jaskolski2007stereochemical}. Thus we define intrinsic geometry at a coarse level, so the resulting signal has more variability. Specifically, we use lengths of edges in the contact graph as a representative of the intrinsic protein geometry, whereas backbone bond orientations capture extrinsic geometry.   

Formally, we model the backbone by the graph, $\gG_b = (\sV_b, \mE_b)$, and the contact graph by the graph, $\gG_t = (\sV_t, \mE_t)$. Then our intrinsic and extrinsic signals, $Int: \mE_t \rightarrow \mathbb{R}$ and $Ext: \mE_b \rightarrow \mathbb{R}^3$ are:
\begin{align}\label{eq:inputs}
    Int(\emE_{ij}) &= \|\emE_{ij}\|_2; \quad \emE_{ij} \in \mE_t, \\
    Ext(\emF_{ij}) &= sgn(j - i) \frac{\emF_{ij}}{\|\emF_{ij}\|}; \quad \emF_{ij} \in \mE_b.
\end{align}

\paragraph{Network Architecture}

With inputs defined, we discuss the architecture of ProGAE. The core idea is to create an \textit{intrinsic} latent space, $L_I \in \mathbb{R}^{n_i}$, and an \textit{extrinsic} latent space, $L_E \in \mathbb{R}^{n_e}$, of dimensions $n_i, n_e$ respectively, via separately encoding the intrinsic and extrinsic signals. Consequently, our network contains two encoders, $Enc_i$ and $Enc_e$ where:
\begin{equation}
    Enc_{i} \circ Int(\mE_t) \in L_I, \qquad Enc_e \circ Ext(\mE_b) \in L_E.     
\end{equation}
We then jointly decode these latent vectors to recover the coordinates of the atoms in the protein backbone. Thus, we formally define the decoder,
    $Dec : L_I \times L_E \rightarrow \mathbb{R}^{|\sV_b| \times 3}$.

This high level structure of ProGAE is depicted in Figure \ref{fig:autoencoder_arch}. We provide additional details on the encoders and decoders. Specific details on layer widths and other parameters can be found in Appendix \ref{subsec:hyperparam}. As the edge-based signals are defined on a geometric domain, it is sensible to learn feature representations using convolutions that respect the geometry of the data. 


As the intrinsic encoder and the extrinsic encoder operate on graphs, the layers of graph attention networks (GATs) introduced in \cite{velivckovic2017graph} are a natural tool to use, albeit with some modification. Since the input signal is defined only on the edges of the graph, $\mE_b$, we define a signal on the graph vertices, $\sV_b$, as the average value of its incident edges, 
\begin{equation}
    f_0(v_i) := \frac{\sum_{j; E_{i, \cdot} \in \mE_b} Ext(\emE_{ij})}{|\{E_{i, \cdot} \in \mE_b\}|} , \quad v_i \in \sV_b.
\end{equation}

Then the first layer of each encoder uses the edge-convolution operator of \cite{gong2019exploiting} to map this edge-defined signal to a vertex-defined signal. The following layers of the extrinsic encoder contains successive graph attention layers with sparsity defined by a given neighborhood radius. At each layer, the signal is downsampled by a factor of two based on farthest point sampling. Given $L$ layers, this defines a sequence of graphs, $\{\gG_{b, i}\}_{i=0}^L$, with increasing decimation. 
Each layer is followed with batch normalization and ReLU. Summarily, for $l=2,...,L$, 
\begin{align}
    & f_l = \sigma \circ BN \circ GAT(d_{l-1}) \text{ s.t. } d_{l-1} = DS(f_{l-1}; 2) \\
    & f_1(v_i) := GAT(f_0(\sV_b), Ext(\mE_b)).
\end{align}
The following layers of the intrinsic encoder are analogous, though we forgo downsampling. 

Global average pooling is applied to the encoder outputs to introduce invariance to size of $\sV_t$ and $\sV_b$. Dense layers then map each result to their respective latent spaces, $L_I$ and $L_E$. The Tanh function is applied to bound the latent space. This produces the intrinsic and extrinsic latent codes, $\vz_i$ and $\vz_e$.  

The latent code $\vz$ is taken as the concatenation of the two latent codes, $[\vz_i, \vz_e]$. A dense layer maps $\vz$ to the a signal defined on the most decimated backbone graph, $\gG_{b, L}$. The structure of the decoder, $Dec$, mirrors $Enc_e$ with convolutions mapping to upscaled graphs. The output of $Dec$ is the point cloud, $\hat{\mP}$, corresponding to the predicted coordinates of the backbone atoms, $\sV_b \approx \mP$.

\paragraph{Loss Function}
\label{subsec:loss}

The loss function is a basic reconstruction loss, where $\mP$ and $\hat{\mP}$ are taken to be the true and predicted coordinates of the protein backbone atoms. Namely, we evaluate their difference using Smooth-$L_1$ loss, $SL_1$. 
This loss, $SL_1(\vx, \vy)$, is defined, with $\delta=2$, as 
\begin{equation}
    \sum_{i=1}^{\#\vx} \min \left(\frac{\delta^2}{2}(\evx_i - \evy_i)^2, \delta |\evx_i - \evy_i| - \frac{1}{2} \right).
\end{equation}
This loss is less sensitive to outliers \cite{girshick2015fast}.

\begin{figure}[tb]
    \centering
    \begin{subfigure}[b]{0.6\linewidth}
        \centering 
        \includegraphics[width=\textwidth]{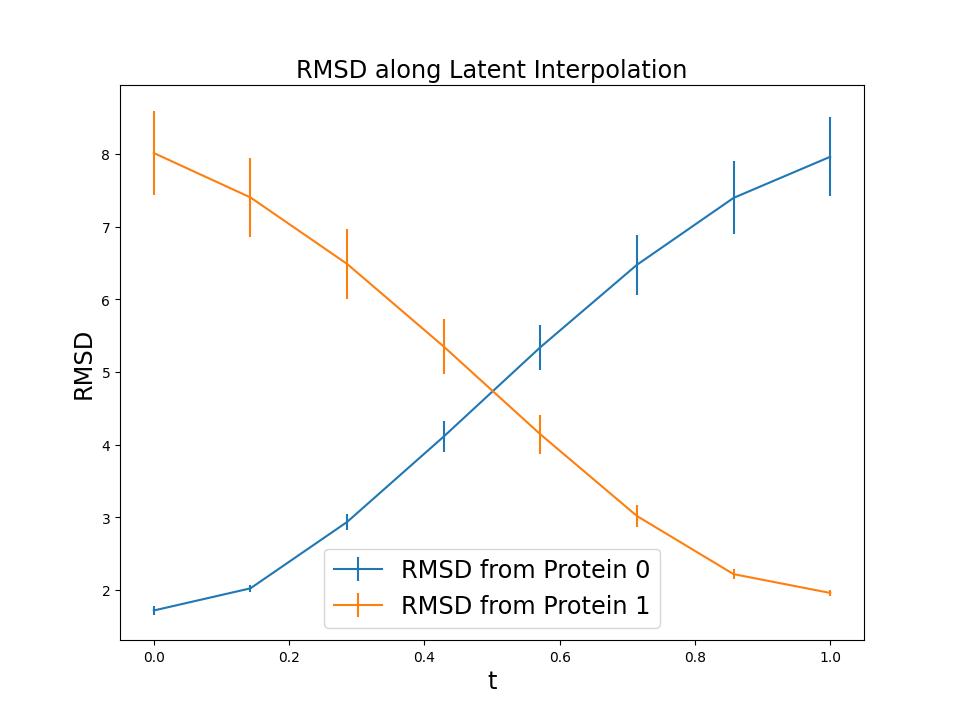}
    \end{subfigure}

    \caption{RMSD of proteins generated along the latent interpolation between two S proteins randomly sampled from different trajectories (see also Figure  \ref{fig:latent_interpolation_hace2} for hACE2 experiments). The RMSDs are computed with respect to the endpoint proteins, with standard error shown. We see a smooth interpolation between the RMSD errors as desired. Examples of the structures along the interpolated path can be found in the appendix in Figure \ref{fig:latent_interpolation_ex}. 
    }
    \label{fig:latent_interpolation}
\end{figure}

    

\section{Experimental Setup}
\label{sec:experimental}
In this section, we describe the setup of our experiments that confirm the usefulness of ProGAE in generating the conformational space. For each dataset, we train three models, each from a different random seed, and report both mean and standard deviation in our results. Our networks are trained on 3 Nvidia GeForce GTX 1080 Ti GPUs. 
\paragraph{Datasets} \label{subsec:datasets}
Datasets used in this work are atomistic simulation trajectories of drug-target proteins reported in \cite{shaw}  
and  \cite{rodriguez2020gpcrmd}. The two main datasets from  \cite{shaw}  used are simulations of proteins in presence of FDA approved or under-investigation molecules, as we aim to test the performance of ProGAE on capturing drug-induced structural variations. This includes the S protein of SARS-CoV-2 and human-ACE2 protein, the former responsible for binding to the later. 
To show that a single ProGAE model can be trained on multiple different proteins, we also utilize  simulations of 38 different G protein-coupled receptors from the GPCRmd dataset \cite{rodriguez2020gpcrmd}. 
More information on these datasets is included in Appendix \ref{app:gpcr_info}.

For comparing with existing work, we run ProGAE on MurD protein simulation data \cite{ramaswamy2019learning} 
. 
For transfer learning, we also consider two trajectories of the entire S protein containing 13,455 backbone atoms from \cite{shaw}. One trajectory is initiated from a closed state, while the other from a partially open state.

\begin{table}[tb]
\centering
\begin{tabular}{@{}rrr@{}}
\toprule
$\textbf{Training loss}$ & S Protein & hACE2 \\
\midrule 
Extrinsic Only & $9.68 \pm 0.09$ E-1 & $4.79 \pm 0.01$ E-1\\
Int. and Ext. & $9.42 \pm 0.08$ E-1 & $4.68 \pm 0.00$ E-1\\
\bottomrule
\end{tabular}
\caption{Training loss on the test sets. We compare models trained on only extrinsic signal and on both intrinsic and extrinsic signals. The inclusion of coarse intrinsic information improves training.   
}
\label{tab:ablation_mse}
\end{table}

\section{Results}

\paragraph{Structure Reconstruction}

\begin{table}[t]
\centering
\begin{tabular}{@{}rrrr@{}}
\toprule
Source data & \textbf{Closed S} &  \textbf{Open S} &  \textbf{Protease} \\
\midrule 
Baseline & $1.55 \pm 0.00$ & $1.76 \pm 0.00$ & $1.14 \pm 0.00$ \\ 
S protein & $1.31 \pm 0.01$ & $\mathbf{1.41 \pm 0.03}$ & $0.96 \pm 0.00$ \\
hACE2 & $\mathbf{1.30 \pm 0.00}$ & $1.42 \pm 0.01$ & $\mathbf{0.93 \pm 0.00}$ \\
\bottomrule
\end{tabular}
\caption{Average atom-wise $L_2$ error in reconstruction (\angstrom) on the test dataset after transferring a trained model to other protein structures. These simulations correspond to two different states of the entire S protein and of the main protease of SARS-CoV-2. 
We retrain only the dense layer mapping to the decoder. 
For comparison, we train the same layer for a randomly initialized model (Baseline). 
}
\label{tab:transfer_results}
\end{table}

Figure \ref{fig:recon} displays the ability of ProGAE to accurately reconstruct conformations. The backbones are visible with atom-wise error in Figures \ref{fig:recon_sarscov2} and \ref{fig:recon_ace2}. 
From the visualized atom-wise $L_2$ reconstruction error, it is clear that our network can capture and reconstruct notable conformational changes of a protein. Figures \ref{fig:recon_sarscov2_noerr} and \ref{fig:recon_ace2_noerr} in the appendix display these reconstructions with color denoting fragment for clarity. In line with the low RMSD error, reconstructed structures appear consistent with ground truths, with larger RMSDs observed in the loop and turn regions. 

Table \ref{tab:reconstruction_results} contains performance metrics of ProGAE. Generalization is measured by the $L_2$ reconstruction error of the backbone atom coordinates, as well as RMSD (root mean square distance) after alignment. For hACE2, we achieve sub-Angstrom performance on the test set. In either case, the RMSD of the reconstruction is within the experimental resolution of the associated PDB files; 6VXX/6VW1 for the S protein and 1R42/1R4L for hACE2. Additionally, the average error in the length of the pseudobonds is also sub-Angstrom. 
Thus, ProGAE is able to reconstruct proteins within meaningful resolution. The RMSD (on secondary structure elements) and $L_2$ error on the benchmark MurD test data  are again lower or comparable to  the experimental resolution and within the range of  what has been reported in the  original study that uses more explicit loss (bond, angle, nonbonded) terms, compared to ProGAE, to preserve protein geometry. 

Additionally, we show in Table \ref{tab:reconstruction_results} that the reconstructed proteins have a large share of their ground-truth contacts recovered. This is measured via the Jaccard index on the binarized contact map using a threshold of $8.0 \angstrom$. Given the set of ground truth contacts, $C_t$, and the set of reconstructed contacts, $C_r$, the Jaccard index is $|C_t \cap C_r| / |C_t \cup C_r|$. Then there is high overlap between the sets of true and ProGAE reconstructed contacts .

\paragraph{Utility of the Extrinsic Latent Space}

With the reconstruction capabilities of ProGAE verified, we consider the benefit of separated intrinsic and extrinsic latent spaces. First, we explore the statistical relationship between the learned intrinsic latent space and the extrinsic latent space. Canonical correlation analysis (CCA) is a natural approach to assess if a linear relationship exists \cite{hardoon2004canonical} (see Appendix \ref{subsub:cca}). 

 Table \ref{tab:cca_results} includes the leading correlation between the intrinsic and extrinsic latent spaces for each dataset, showing 
that the correlation between intrinsic and extrinsic latent space is non-negligible. However, we find that the extrinsic latent space is much more linearly separable concerning conformational properties reflective of specific drug binding. 
As stated earlier, each trajectory in the dataset corresponds to the S or hACE2 protein bound to a specific drug. Then it is natural to investigate if this distinct drug information is encoded in the two latent spaces. Table \ref{tab:cca_results} contains the performance of a linear classifier trained via one-shot learning on the different latent spaces to classify the drug present in  each frame. 
It is clear that the drug molecule  can be almost perfectly classified in the extrinsic latent space, while such classification is much weaker in the intrinsic latent space. Figures \ref{fig:cca_sarscov2} and \ref{fig:cca_ace2} display the embeddings of the test set in the latent spaces, projected to the first two canonical components. Color denotes the identity of the drug that the protein is bound to. Even in the 2D projection of the extrinsic latent space, better clustering by the drug identity is apparent. 

To check  if this linear separation is chemically meaningful, we train a linear regression model on the extrinsic latent space to predict physico-chemical properties of a drug binding to a protein. 
Table \ref{tab:classifier_results} displays the performance of the model at predicting the properties of molecular weight, hydrogen bond donor count, and topological polar surface area. For comparison to our latent embedding, we train a linear regression model on the first $n_e$ principal component scores of the PCA of the extrinsic signal on each element of the test dataset. The latent regression outperforms that of PCA, indicating that  the extrinsic latent embedding captures more  physico-chemical information about the bound drug. We believe this linear regression is appropriate to prevent overfitting.

\paragraph{Utility of the Intrinsic Latent Space}

We now weigh the benefits of including the intrinsic latent space in the model. 
We find the inclusion of the intrinsic latent space improves the performance of the learned network during training. To see this, we trained a model that only encodes the extrinsic signal to reconstruct the protein. While it was comparable in performance regarding $L_2$ error, we found this extrinsic-only model resulted in a higher training loss on the test set.
This is shown in Table \ref{tab:ablation_mse}. 

\paragraph{Geometry and Steerability of the Disentangled Latent Space}


 For a generative model, it is important to consider if paths in the latent space are smooth. 
 We evaluate the performance of linear interpolations in the learned latent space. Given two protein conformations from different trajectories (i.e. in the context of two different drugs), we generate a path between them by generating the linear interpolation of their latent codes. This provides a path of structural variation that does not exist in the training data. The results of this interpolation in terms of RMSD is shown in Figure \ref{fig:latent_interpolation}. A smooth exchange in the RMSD error of the generated protein structures from the two endpoints is evident. 
 

\vspace{-0.5em}
\paragraph{Transfer Learning-- Extension to Different Proteins}
\label{subsec:transfer}

To check the generalization of ProGAE, we investigate transfer learning to simulations of  different proteins. We begin with models trained on the S protein comprised of the 3 RBDs  and on  hACE2. These results are summarized in Table \ref{tab:transfer_results}. We transfer learned ProGAE models to  trajectories of the closed and partially open state of the entire S protein, as well as SARS-COV-2 main Protease, which provides insight into the generalization capability of the learned convolutional filters. 
As a result, six scenarios of the transfer learning, in addition to three random baselines, are reported in Table \ref{tab:transfer_results}. When transferring the model trained on the 3 RBDs of the S protein to the S protein in the closed  state, we are transferring the model learned on a partial structure to the entire protein \textit{that is much larger in size}. Model transfer to the S protein in the partially open state deals with a scenario where \textit{the conformational state of the protein is notably different (closed vs partially open)}. Transferring the model trained on hACE2 to the S protein datasets studies the knowledge transfer to an \textit{entirely different protein, but one which hACE2 is known to interact with}. Finally, transferring both S protein and hACE2 models to the main Protease simulation allows us to study the transfer to a \textit{completely different protein without notable interaction with the source protein.} As performing long time-scale simulations of large protein systems at high resolution is computationally expensive, our method appears beneficial, as ProGAE  transfers well to non-related proteins of larger size. 

The only incompatible layer is the dense layer mapping from the latent spaces.
To investigate transfer learning, we train just this dense layer for 10 epochs. As a baseline, we train the same layer of a randomly initialized model. In all cases, the  transferred model performs better than the baseline. Thus the learned filters generalize to trajectories of different protein systems. Results on transferring only intrinsic/extrinsic filters are in the appendix in Table \ref{tab:transfer_int_ext}.

\section{Conclusion}

We introduce a novel geometric autoencoder, ProGAE, for learning meaningful disentangled representations of the protein conformational space. Our model accurately reconstructs  structures of established drug-target proteins such as SARS-CoV-2 Spike, human ACE2, and GPCR proteins, as well as an existing  benchmark MurD. 
The autoencoder separately encodes intrinsic and extrinsic geometries to ensure better latent interpretability. The extrinsic latent space can classify structures with respect to their bound drug molecules, as well as can predict the drug properties. The intrinsic space assists in improving the quality of the reconstructions. The resulting disentangled, smooth latent space enables controllable generation of protein structures in a drug-dependent manner. We also show that the filters learned in training can be successfully transferred to trajectories of different protein systems, irrespective of system size, conformational state, or presence of protein-protein interaction. These results on learning, prediction, and generation suggest that the proposed framework  can serve as a step towards bridging geometric deep learning with protein simulations and provide an efficient and complimentary means for understanding and enhancing  structural landscapes of important drug-target proteins. 

\section*{Acknowledgments}
J. Tatro's work was supported by the IBM-RPI AIRC program. R. Lai's work is supported in part by NSF CAREER Award (DMS—1752934)





\begin{small}
\bibliographystyle{ieeetr}
\bibliography{main}

\begin{thebibliography}{10}

\bibitem{jumper2021highly}
J.~Jumper, R.~Evans, A.~Pritzel, T.~Green, M.~Figurnov, O.~Ronneberger,
  K.~Tunyasuvunakool, R.~Bates, A.~{\v{Z}}{\'\i}dek, A.~Potapenko, {\em
  et~al.}, ``Highly accurate protein structure prediction with alphafold,''
  {\em Nature}, vol.~596, no.~7873, pp.~583--589, 2021.

\bibitem{ramaswamy2020learning}
V.~K. Ramaswamy, C.~G. Willcocks, and M.~T. Degiacomi, ``Learning protein
  conformational space by enforcing physics with convolutions and latent
  interpolations,'' 2020.

\bibitem{bhowmik2018deep}
D.~Bhowmik, S.~Gao, M.~T. Young, and A.~Ramanathan, ``Deep clustering of
  protein folding simulations,'' {\em BMC bioinformatics}, vol.~19, no.~18,
  pp.~47--58, 2018.

\bibitem{guo2020generating}
X.~Guo, S.~Tadepalli, L.~Zhao, and A.~Shehu, ``Generating tertiary protein
  structures via an interpretative variational autoencoder,'' {\em arXiv
  preprint arXiv:2004.07119}, 2020.

\bibitem{varolgunecs2020interpretable}
Y.~B. Varolg{\"u}ne{\c{s}}, T.~Bereau, and J.~F. Rudzinski, ``Interpretable
  embeddings from molecular simulations using gaussian mixture variational
  autoencoders,'' {\em Machine Learning: Science and Technology}, vol.~1,
  no.~1, p.~015012, 2020.

\bibitem{tatro2020unsupervised}
N.~J. Tatro, S.~C. Schonsheck, and R.~Lai, ``Unsupervised geometric
  disentanglement for surfaces via cfan-vae,'' 2020.

\bibitem{wu2019disentangling}
W.~Wu, K.~Cao, C.~Li, C.~Qian, and C.~C. Loy, ``Disentangling content and style
  via unsupervised geometry distillation,'' 2019.

\bibitem{yang2020dsmnet}
J.~Yang, K.~Mo, Y.-K. Lai, L.~J. Guibas, and L.~Gao, ``Dsm-net: Disentangled
  structured mesh net for controllable generation of fine geometry,'' 2020.

\bibitem{graves2020review}
J.~Graves, J.~Byerly, E.~Priego, N.~Makkapati, S.~V. Parish, B.~Medellin, and
  M.~Berrondo, ``A review of deep learning methods for antibodies,'' {\em
  Antibodies}, vol.~9, no.~2, p.~12, 2020.

\bibitem{jing2020learning}
B.~Jing, S.~Eismann, P.~Suriana, R.~J.~L. Townshend, and R.~Dror, ``Learning
  from protein structure with geometric vector perceptrons,'' 2020.

\bibitem{gainza2019deciphering}
P.~Gainza, F.~Sverrisson, F.~Monti, E.~Rodola, M.~Bronstein, and B.~Correia,
  ``Deciphering interaction fingerprints from protein molecular surfaces,''
  {\em bioRxiv}, p.~606202, 2019.

\bibitem{hermosilla2020proteinn}
P.~Hermosilla, M.~Schäfer, M.~Lang, G.~Fackelmann, P.~P. Vázquez,
  B.~Kozlíková, M.~Krone, T.~Ritschel, and T.~Ropinski, ``Proteinn:
  Intrinsic-extrinsic convolution and pooling for scalable deep protein
  analysis,'' 2020.

\bibitem{ramaswamy2019learning}
V.~K. Ramaswamy, C.~G. Willcocks, and M.~T. Degiacomi, ``Learning protein
  conformational space by enforcing physics with convolutions and latent
  interpolations,'' 2019.

\bibitem{vendruscolo1997recovery}
M.~Vendruscolo, E.~Kussell, and E.~Domany, ``Recovery of protein structure from
  contact maps,'' {\em Folding and Design}, vol.~2, no.~5, pp.~295--306, 1997.

\bibitem{jaskolski2007stereochemical}
M.~Jaskolski, M.~Gilski, Z.~Dauter, and A.~Wlodawer, ``Stereochemical
  restraints revisited: how accurate are refinement targets and how much should
  protein structures be allowed to deviate from them?,'' {\em Acta
  Crystallographica Section D: Biological Crystallography}, vol.~63, no.~5,
  pp.~611--620, 2007.

\bibitem{velivckovic2017graph}
P.~Veli{\v{c}}kovi{\'c}, G.~Cucurull, A.~Casanova, A.~Romero, P.~Lio, and
  Y.~Bengio, ``Graph attention networks,'' {\em arXiv preprint
  arXiv:1710.10903}, 2017.

\bibitem{gong2019exploiting}
L.~Gong and Q.~Cheng, ``Exploiting edge features for graph neural networks,''
  in {\em Proceedings of the IEEE Conference on Computer Vision and Pattern
  Recognition}, pp.~9211--9219, 2019.

\bibitem{girshick2015fast}
R.~Girshick, ``Fast r-cnn,'' in {\em Proceedings of the IEEE international
  conference on computer vision}, pp.~1440--1448, 2015.

\bibitem{shaw}
{D.E. Shaw Research}, ``Molecular dynamics simulations related to sars-cov-2.''
  \url{ http://www.deshawresearch.com/resources_sarscov2.html}, 2020.
\newblock Accessed: 2020-09-30.

\bibitem{rodriguez2020gpcrmd}
I.~Rodr{\'\i}guez-Espigares, M.~Torrens-Fontanals, J.~K. Tiemann,
  D.~Aranda-Garc{\'\i}a, J.~M. Ram{\'\i}rez-Anguita, T.~M. Stepniewski,
  N.~Worp, A.~Varela-Rial, A.~Morales-Pastor, B.~Medel-Lacruz, {\em et~al.},
  ``Gpcrmd uncovers the dynamics of the 3d-gpcrome,'' {\em Nature Methods},
  vol.~17, no.~8, pp.~777--787, 2020.

\bibitem{hardoon2004canonical}
D.~R. Hardoon, S.~Szedmak, and J.~Shawe-Taylor, ``Canonical correlation
  analysis: An overview with application to learning methods,'' {\em Neural
  computation}, vol.~16, no.~12, pp.~2639--2664, 2004.

\bibitem{wang2020differentiable}
W.~Wang, S.~Axelrod, and R.~Gómez-Bombarelli, ``Differentiable molecular
  simulations for control and learning,'' 2020.

\bibitem{chen2020symplectic}
Z.~Chen, J.~Zhang, M.~Arjovsky, and L.~Bottou, ``Symplectic recurrent neural
  networks,'' 2020.

\bibitem{ingraham2019learning}
J.~Ingraham, A.~J. Riesselman, C.~Sander, and D.~S. Marks, ``Learning protein
  structure with a differentiable simulator.,'' in {\em ICLR}, 2019.

\bibitem{Mardt_2018}
A.~Mardt, L.~Pasquali, H.~Wu, and F.~Noé, ``Vampnets for deep learning of
  molecular kinetics,'' {\em Nature Communications}, vol.~9, Jan 2018.

\bibitem{lee2019deepdrivemd}
H.~Lee, H.~Ma, M.~Turilli, D.~Bhowmik, S.~Jha, and A.~Ramanathan,
  ``Deepdrivemd: Deep-learning driven adaptive molecular simulations for
  protein folding,'' 2019.

\bibitem{tsai2020learning}
S.-T. Tsai, E.-J. Kuo, and P.~Tiwary, ``Learning molecular dynamics with simple
  language model built upon long short-term memory neural network,'' 2020.

\bibitem{chen2018molecular}
W.~Chen and A.~L. Ferguson, ``Molecular enhanced sampling with autoencoders:
  On-the-fly collective variable discovery and accelerated free energy
  landscape exploration,'' {\em Journal of computational chemistry}, vol.~39,
  no.~25, pp.~2079--2102, 2018.

\bibitem{chen2019nonlinear}
W.~Chen, H.~Sidky, and A.~L. Ferguson, ``Nonlinear discovery of slow molecular
  modes using state-free reversible vampnets,'' {\em The Journal of chemical
  physics}, vol.~150, no.~21, p.~214114, 2019.

\bibitem{ribeiro2018reweighted}
J.~M.~L. Ribeiro, P.~Bravo, Y.~Wang, and P.~Tiwary, ``Reweighted autoencoded
  variational bayes for enhanced sampling (rave),'' {\em The Journal of
  chemical physics}, vol.~149, no.~7, p.~072301, 2018.

\bibitem{bonati2019neural}
L.~Bonati, Y.-Y. Zhang, and M.~Parrinello, ``Neural networks-based
  variationally enhanced sampling,'' {\em Proceedings of the National Academy
  of Sciences}, vol.~116, no.~36, pp.~17641--17647, 2019.

\bibitem{zhang2019targeted}
J.~Zhang, Y.~I. Yang, and F.~No{\'e}, ``Targeted adversarial learning optimized
  sampling,'' {\em The journal of physical chemistry letters}, vol.~10, no.~19,
  pp.~5791--5797, 2019.

\bibitem{noe2019boltzmann}
F.~No{\'e}, S.~Olsson, J.~K{\"o}hler, and H.~Wu, ``Boltzmann generators:
  Sampling equilibrium states of many-body systems with deep learning,'' {\em
  Science}, vol.~365, no.~6457, p.~eaaw1147, 2019.

\bibitem{kingma2014adam}
D.~P. Kingma and J.~Ba, ``Adam: A method for stochastic optimization,'' {\em
  arXiv preprint arXiv:1412.6980}, 2014.

\end{thebibliography}
\end{small}

\clearpage


\appendix

\section{Appendix}

\begin{figure}[t]
\centering
 \begin{subfigure}[b]{0.45\textwidth}
     \centering
     \includegraphics[width=\textwidth]{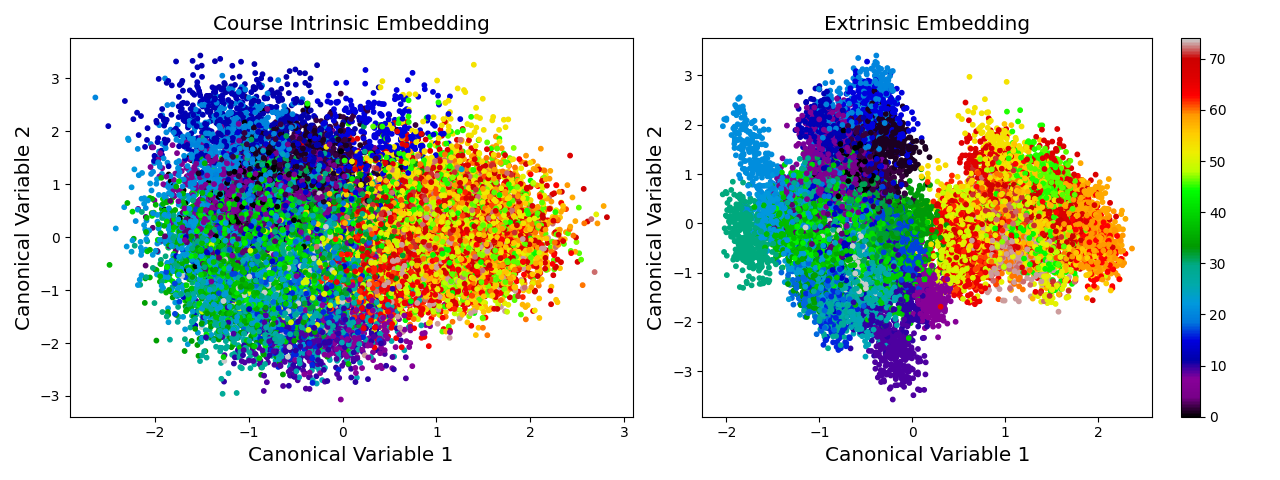}
     \caption{Human ACE2}
     \label{fig:cca_ace2}
 \end{subfigure}
\caption{Displays drug clustering as in Figure \ref{fig:cca_sarscov2}, but for hACE2.}
\end{figure}

\begin{figure}[tb]
\centering
\begin{subfigure}[b]{0.45\textwidth}
     \centering
     \includegraphics[width=\textwidth]{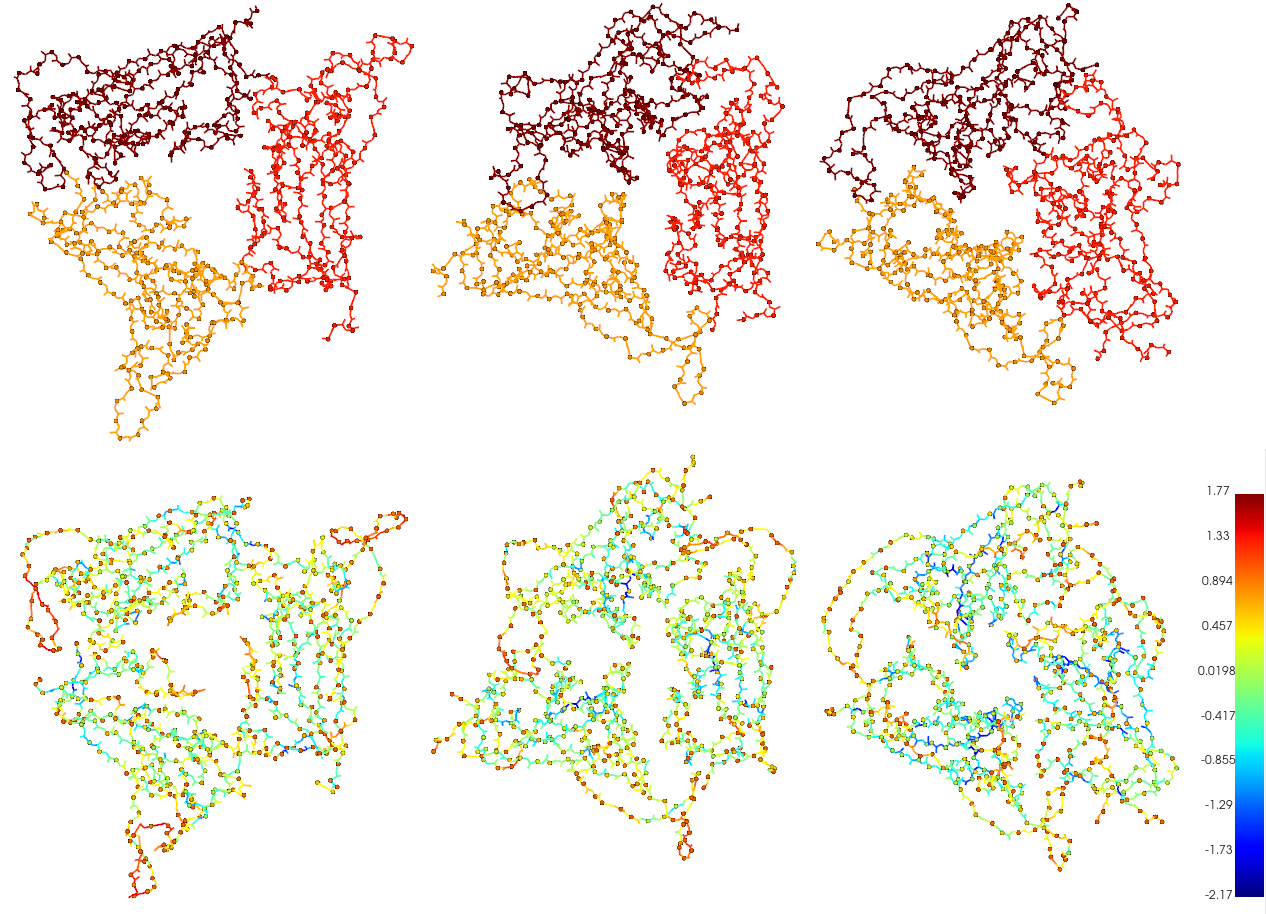}
     \caption{3 RBDs of S protein}
     \label{fig:recon_sarscov2}
 \end{subfigure}
 \begin{subfigure}[b]{0.45\textwidth}
     \centering
     \includegraphics[width=\textwidth]{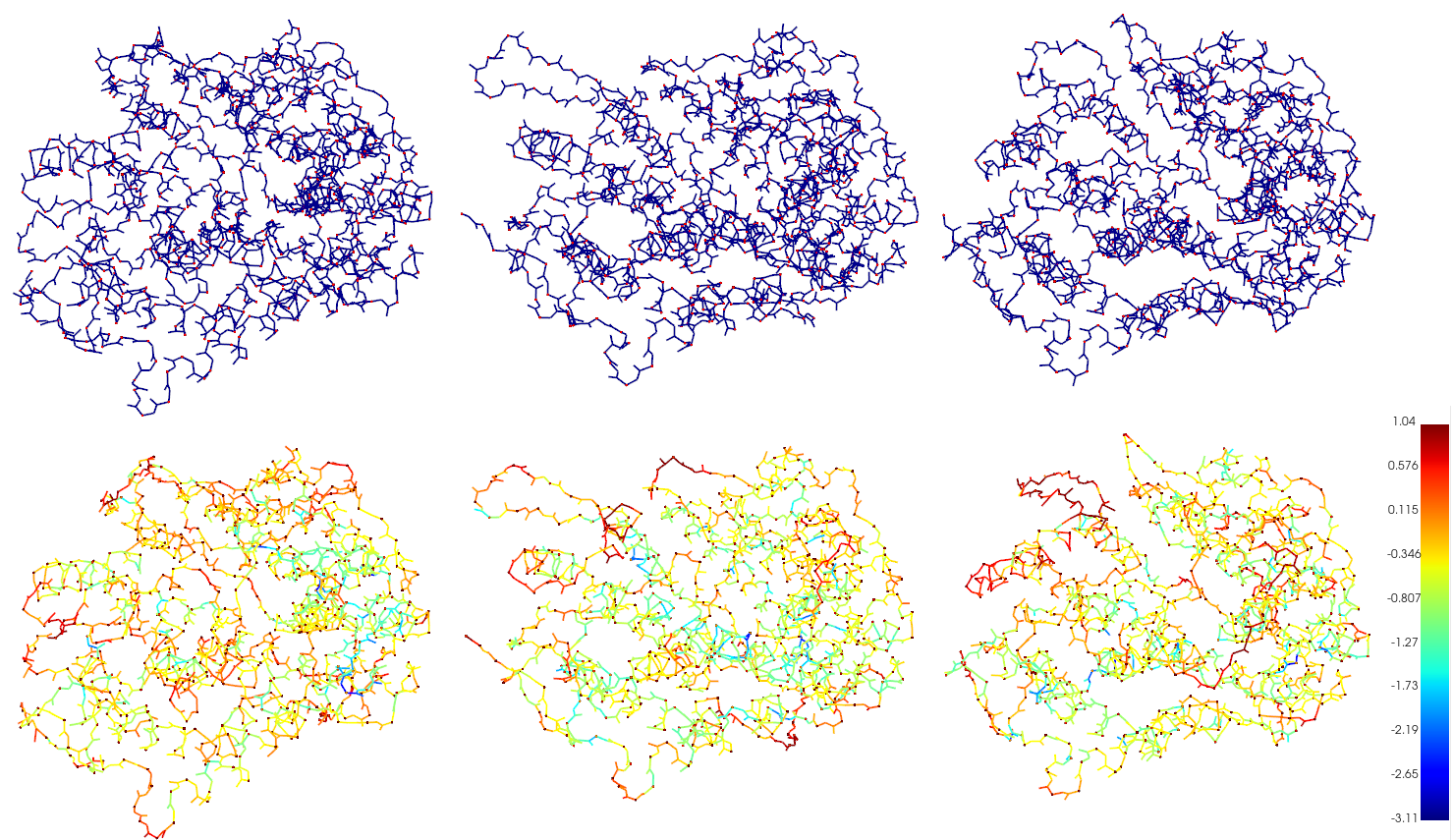}
     \caption{Ectodomain of Human ACE2}
     \label{fig:recon_ace2}
 \end{subfigure}
\caption{Reconstructions of protein frames from test data using ProGAE. The top row displays the ground truth, while the bottom row displays the corresponding structure generated by the network.  
Color in the top row denotes separate protein chains, while color in the bottom row indicates the log of  atom-wise $L_2$ error. Color of the bonds indicates the average of the constituent atoms. }
\end{figure}


\subsection{Additional Background}
\label{app:add_back}

\paragraph{Related Work on Modeling Dynamics} For completeness, we discuss recent work  modeling temporal aspects of  protein dynamics here. \cite{wang2020differentiable} learns the dynamics between a particular starting conformation and a particular target conformation by training a control Hamiltonian represented by a graph neural network. From the perspective of Hamiltonian systems, \cite{chen2020symplectic} introduces Symplectic RNNs, which leverages sympletic integration for learning the dynamics of a physical system. \cite{ingraham2019learning} learns dynamics via a differentiable simulator using Langevin dynamics. 

\cite{Mardt_2018} introduces VAMPnet, which maps molecular coordinates to Markov states, to capture molecular kinetics.  \cite{lee2019deepdrivemd} augments molecular dynamics simulation with deep learning to improve the sampling of the folded states of proteins.  \cite{tsai2020learning} uses a LSTM to predict protein dynamics.  

Another line of work involves enhanced sampling with neural net approximations of slow variables  \cite{chen2018molecular, chen2019nonlinear, ribeiro2018reweighted}. Deep and adversarial learning have been employed for such enhanced sampling \cite{bonati2019neural, zhang2019targeted}. \cite{noe2019boltzmann} uses a deep generative neural net to directly sample the equilibrium distribution of a many-body system defined by an energy function, without using molecular simulation.

\subsection{Canonical Correlation Analysis} \label{subsub:cca}
Given the two embeddings, $\mX \in \mathbb{R}^{n \times m_1}$ and $\mY \in \mathbb{R}^{n \times m_2}$, CCA finds two linear transformations, $\mA \in \mathbb{R}^{m_1 \times m_1}$ and $\mB \in \mathbb{R}^{m_2 \times m_2}$, such that
\begin{align}\label{eq:cca}
    & \mA, \mB = \argmin_{\mA, \mB} ||\mX \mA - \mY \mB||_F^2 \\
    & \quad \text{s.t.} \quad \mX^T \mA^T \mA \mX = \mI_{m_1}, \mY^T \mB^T \mB \mY = \mI_{m_2}.
\end{align}
\Eqref{eq:cca} has a closed form solution given by SVD. It follows that $\max{}diag(\mX^T \mA^T \mB \mY)$ denotes the highest correlation between any axis in each embedding.

\begin{table*}[t]
\centering
\begin{tabular}{@{}rrcrcr@{}}
\toprule
Source dataset & \textbf{Closed S} &  & \textbf{Partially Open S} &  & \textbf{Protease} \\
\midrule 
S protein (\angstrom) \\
\cmidrule{1-1} 
Intrinsic only & $1.30 \pm 0.00$ && $1.47 \pm 0.00$ && $0.97 \pm 0.00$ \\
Extrinsic only & $1.32 \pm 0.01$ && $1.41 \pm 0.03$ && $0.98 \pm 0.00$\\
hACE2 (\angstrom) \\
\cmidrule{1-1} 
Intrinsic only & $1.30 \pm 0.01$ && $1.47 \pm 0.01$ && $0.96 \pm 0.00$ \\
Extrinsic only & $1.30 \pm 0.00$ && $1.43 \pm 0.01$ && $0.95 \pm 0.00$\\
\bottomrule
\end{tabular}
\caption{Average atom-wise $L_2$ error in reconstruction (\angstrom) on the test dataset after transferring a trained model to other protein structures. Analogous to Table \ref{tab:transfer_results}, though we only transfer either the intrinsic or extrinsic convolutional filters. The results outperform the baseline from Table \ref{tab:transfer_results}, and both types of convolutional filters are seen to be valuable.}
\label{tab:transfer_int_ext}
\end{table*}

\subsection{Network Hyperparameters}
\label{subsec:hyperparam}

Here we specify the hyperparameters of the networks used in conducting our experiments. Each encoder contains 5 layers with filter sizes, $\{12, 24, 48, 96, 96\}$. The decoder structure is mirrored with filter sizes, $\{128, 128, 64, 32, 16, 3\}$. Each graph attention layer has 4 heads of attention. The dimensions of the intrinsic and extrinsic latent spaces are set to 16 and 32 respectively.

For training, we use ADAM with a learning rate of 1E-3 \cite{kingma2014adam}. Learning rate decays at a rate of 0.995 per epoch. We train models with a weight decay penalty of 5E-5. The models are trained 100 epochs, which is enough to achieve convergence, with a batch size of 64. Additionally, we set $\lambda_{\mathcal{R}} = 5$E-1 for the bond length penalty.
The neighborhood radius for defining the sparsity of the graph attention layer is set to 2.5 $\angstrom$ in the first layer. This radius is scaled at each layer with the stride of the previous convolution.


\begin{figure}[tb]
    \centering
    \begin{subfigure}[b]{0.85\linewidth}
        \centering 
        \includegraphics[width=\textwidth]{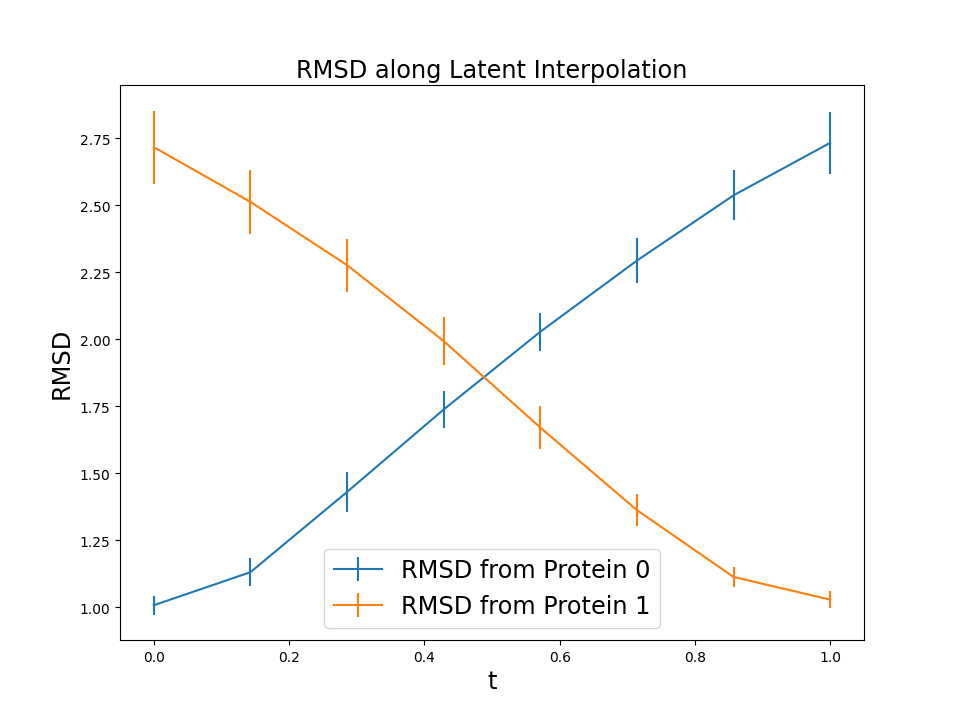}
        \caption{Human ACE2}
    \end{subfigure}
    
    \caption{RMSD of proteins generated along the latent interpolation between two hACE2 proteins randomly sampled from different trajectories. The RMSDs are computed with respect to the endpoint proteins, with standard error shown. We see a smooth interpolation between the RMSD errors as desired.}
    \label{fig:latent_interpolation_hace2}
\end{figure}

\begin{figure*}[tbh]
    \centering
    \includegraphics[width=\textwidth]{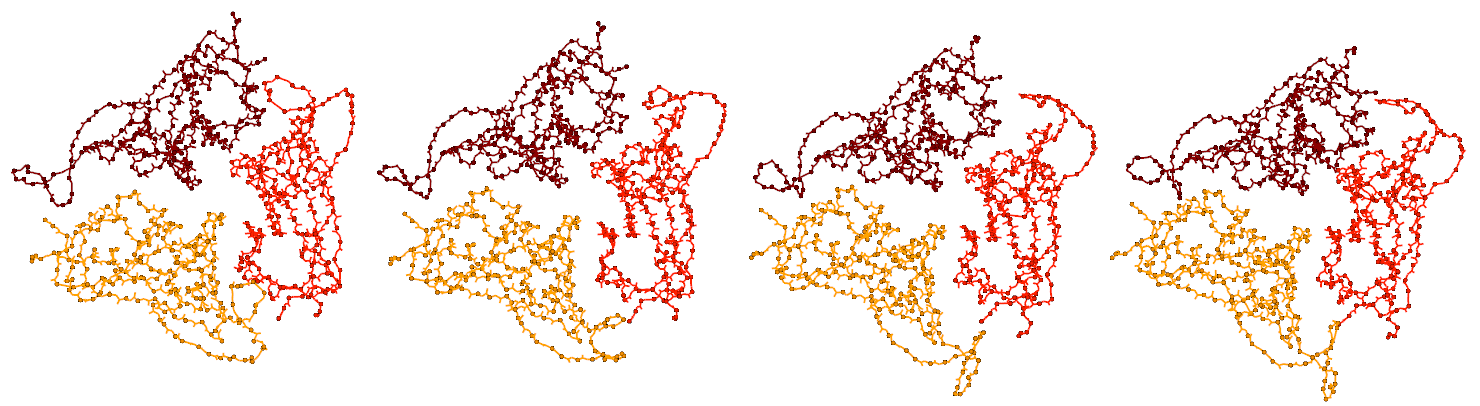}
    \caption{A sample latent interpolation between S proteins from different trajectories. This shows the smooth transition along the latent path analyzed in Figure \ref{fig:latent_interpolation}.}
    \label{fig:latent_interpolation_ex}
\end{figure*}

\begin{figure*}[tbh]
\centering
\begin{subfigure}[b]{0.9\textwidth}
     \centering
     \includegraphics[width=\textwidth]{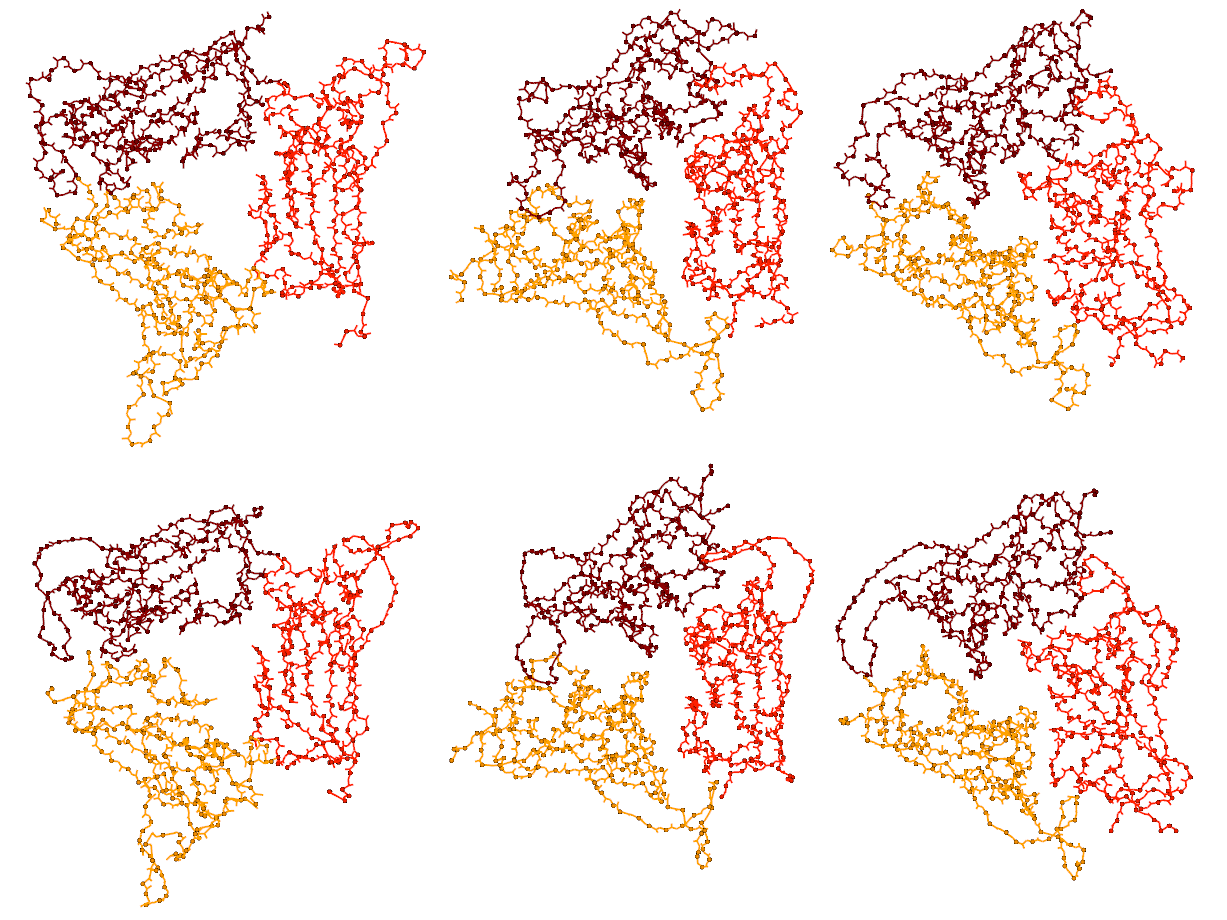}
     \caption{3 RBDs of S protein}
     \label{fig:recon_sarscov2_noerr}
 \end{subfigure}
 \begin{subfigure}[b]{0.9\textwidth}
     \centering
     \includegraphics[width=\textwidth]{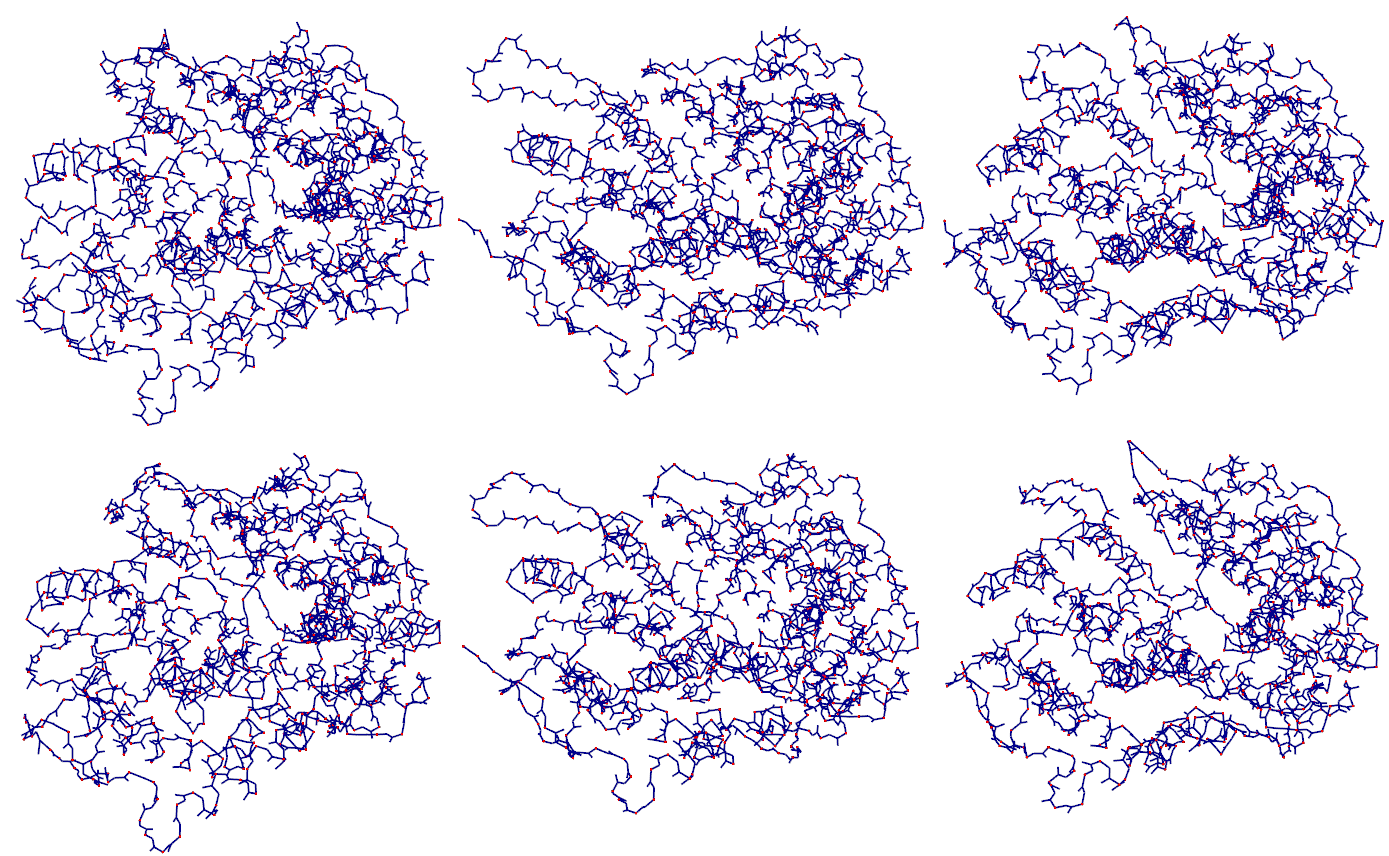}
     \caption{Ectodomain of Human ACE2}
     \label{fig:recon_ace2_noerr}
 \end{subfigure}
\caption{The protein reconstructions from Figures \ref{fig:recon_sarscov2} and \ref{fig:recon_ace2}. The color in the bottom row indicates different protein fragments, instead of the log of atom-wise $L_2$ error. 
}
\end{figure*}


\subsection{Additional Details on Datasets}
\label{app:gpcr_info}
Here we include additional details on the datasets used for experiments. 
These datasets in \cite{shaw} are:
(1) \textit{50 independent trajectories}, each simulating the SARS-CoV-2 trimeric spike protein (S protein) in the presence of a distinct drug for 2$\mu$s. The simulation is limited to 3 receptor binding domains (RBDs) of the protein, as well as a short region needed for the system to maintain a trimer assembly; 
(2) \textit{75 independent trajectories}, each simulating the ectodomain protein of human ACE2 (hACE2) in the presence of a distinct drug for 2$\mu$s. 

The backbones of the S protein and the hACE2 protein contain 3,690 atoms and 2,386 atoms, respectively. The time resolution is 1,200 ps. We form the training, validation, and test sets via randomly sampling frames with a 70\%/10\%/20\% split.

To form our GPCRmd datasets for training and testing our network, we sample 38 simulations of different proteins from this dataset. These proteins are G-protein coupled receptors. This includes simulations of various different Rhodopsin and Secretin proteins. 
The IDs of the simulations used are 21, 35, 59, 61, 63, 66, 68, 72, 73, 75, 77, 82, 85, 87, 91, 92, 93, 95, 98, 105, 108, 110, 111, 118, 122, 128, 141, 154, 155, 157, 158, 163, 166, 171, 175, 179, 186, 216.

Regarding the transfer learning dataset, we use the first 2.5 $\mu s$ of these 10 $\mu s$ simulations, corresponding to 2,001 frames with a resolution of 1,200 ps. Additionally, we utilize the first 10 $\mu s$ of a 100 $\mu s$ simulation of the main Protease of SARS-CoV-2, a sequence of 10,001 frames with a 1,000 ps resolution.
\end{document}